\documentclass{revtex4}

\usepackage{latexsym}
\usepackage{graphics}
\usepackage{amsmath}
\usepackage{amssymb}
\usepackage{gensymb}
\usepackage[usenames,dvipsnames]{color}
\usepackage{epstopdf}
\usepackage{bm}
\usepackage{amsthm}
\usepackage{graphicx}

\usepackage[T1]{fontenc}
\usepackage{fourier}
\usepackage{subcaption}
\usepackage{dsfont}
\usepackage{amsthm}
\usepackage[utf8]{inputenc}

\begin{document}
\title{Monte Carlo Tree Search for a single target search game on a 2-D lattice}

\author{Elana Kozak}
\affiliation{Department of Mathematics, United States Naval Academy, Annapolis, Maryland}
\author{Scott Hottovy}
\affiliation{Department of Mathematics, United States Naval Academy, Annapolis, Maryland}



\begin{abstract}
Monte Carlo Tree Search (MCTS) is a branch of stochastic modeling that utilizes decision trees for optimization, mostly applied to artificial intelligence (AI) game players. This project imagines a ?game? in which an AI player searches for a stationary target within a 2-D lattice. We analyze its behavior with different target distributions and compare its efficiency to the Levy Flight Search, a model for animal foraging behavior. In addition to simulated data analysis we prove two theorems about the convergence of MCTS when computation constraints disappear.
\end{abstract}

\maketitle

\section{Introduction}

The problem of search and detection is applicable in many scenarios, both theoretical and practical. The concept of search and detection is complex, with many possible variations and applications. For example, a simple search could include one search agent and one target, both randomly moving throughout a space \cite{gage94}. Many methods for finding these targets have been tested and applied with varying levels of success \cite{bellman}. We propose a new method to solve the problem of search and detection: Monte Carlo Tree Search (MCTS).

To test the MCTS method we start with a simple 2-D lattice search problem. The domain is a $N \times N$ lattice with periodic boundaries and discrete locations for the target and searcher to occupy. That is, the searcher at time $t$ has position 
\begin{equation}
	S_t = (x,y), \mbox{ such that }x,y\in\{1,2,3,...,N\}. 
\end{equation}
Further more
The target $T$ is placed within the search domain according to some desired distribution $P(x,y)$ on the lattice. The searcher starts at a predetermined location, typically $(1,1)$. 

While playing the "game" the searcher can make moves in one of four directions: up, down, left, or right. That is, if $S_t = (x,y)$ then 
\begin{equation}
	S_{t+1} = S_t + \xi, 
\end{equation} 
where $\xi \in\{(0,1),(0,-1),(1,0),(-1,0)\}$. All steps are one unit in length and the boundaries are periodic, i.e. if $S_t = (x,N)$ and $\xi = (0,1)$ then $S_{t+1} = (x,1)$. Similarly for the other boundaries. The target is stationary. 

\begin{figure}[h] 
\centering
  \begin{subfigure}[b]{0.4\linewidth}
    \centering
    \includegraphics[width=.9\linewidth]{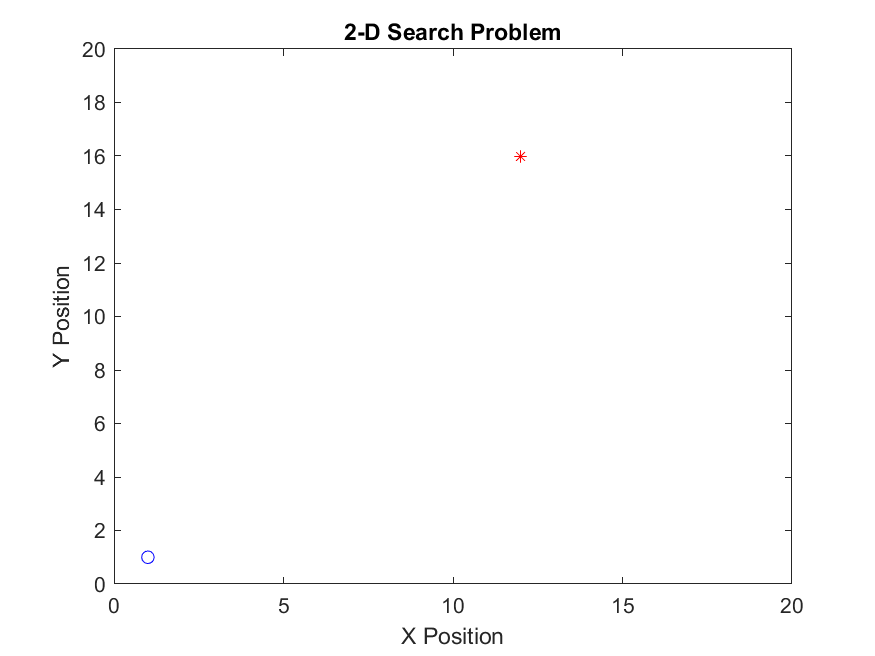} 
    \label{fig1:a} 
  \end{subfigure}
  \begin{subfigure}[b]{0.4\linewidth}
    \centering
    \includegraphics[width=.9\linewidth]{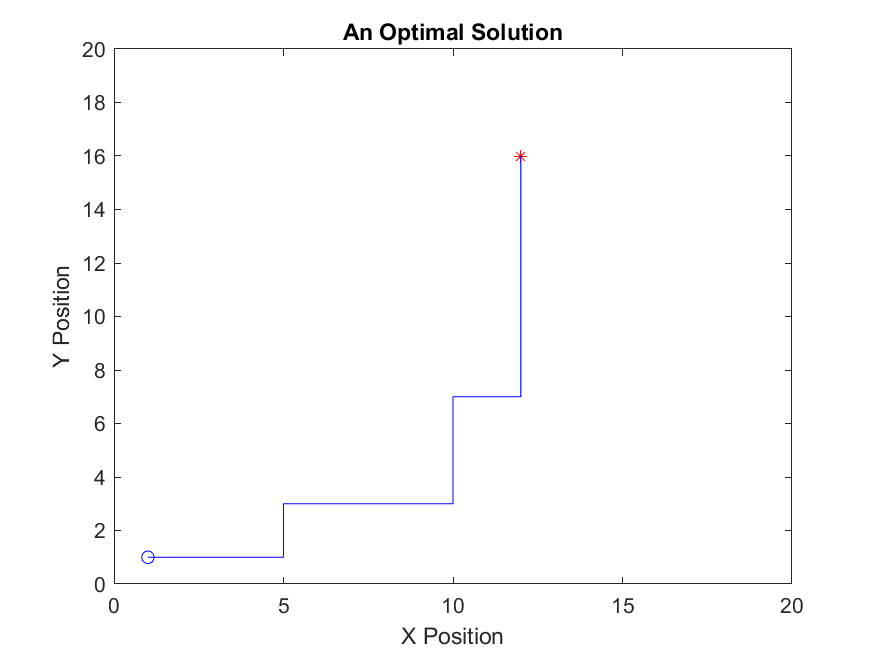} 
    \label{fig1:b} 
  \end{subfigure} 
\caption{Example 2-D search problem and one of its optimal solutions.}
  \label{fig1} 
\end{figure}

The solution to this search problem is a path to the target which can be measured by the number of steps. An optimal path is defined a set of moves which reaches the target in the minimum number of steps. This requires the searcher to always move directly towards the target in either the horizontal or vertical directions. A detection occurs when the searcher occupies the target's position $S_\tau = T$ for some time $\tau$. Achieving this optimal path is the best solution to this search problem. An example problem and one optimal path is shown in figure \ref{fig1}. 

The main focus of this paper is a method called the Monte Carlo Tree Search (MCTS), defined as "a method for finding optimal decisions in a given domain by taking random samples in the decision space and building a search tree according to the results" \cite{BrowneSurvey}. The method has so far been used to develop artificial intelligence game players for games such as Go and chess. Important advantages include the ability to set a decision time limit and the lack of domain knowledge needed for success \cite{TreeSearchBoardGame}. While very useful in certain board games, the process is at its core a decision making method, allowing us to apply the theoretical concepts to any choice, including search paths.

The Monte Carlo Tree Search depends on a game tree which consists of many nodes that represent possible game states. For our ``game'' of search and detection, each searcher location is a game state. The MCTS method works by running many simulations to learn about the game tree in order to choose its next move. The general process includes four steps: selection, expansion, playout, and back-propagation. Most significantly, the selection policy we use is the Upper Confidence Bound for Trees (UCT), which chooses the node with maximum UCT value defined by 
\begin{equation}
UCT(v_i, v) = \frac{Q(v_i)}{N(v_i)} + c \sqrt{\frac{log(N(v))}{N(v_i)}}.
\end{equation}
Here, $v$ represents the root node, $v_i$ represents a child node, $Q(v)$ is the average reward, $N(v)$ is the number of visits, and $c$ is a variable constant that balances exploration and exploitation \cite{ComparisonMCTS}. For this study we choose $c = \sqrt{2}$ as it has been shown to be the optimal exploration coefficient for similar problems CITE\cite{...}. The reward function $Q(v_i)$ is chosen as
	$$ Q(v_i) = \frac{1}{\tau_{v_i,N}}$$  
where $\tau_{v_i,N}$ is the stopping time for the default policy to find the target for the first time starting from node $v_i$ 
on an $N\times N$ grid. Thus a shorter time results in a higher reward. 

\begin{figure}[htbp] 
\centering
  \begin{subfigure}[b]{0.45\linewidth}
    \centering
    \includegraphics[width=1.0\linewidth]{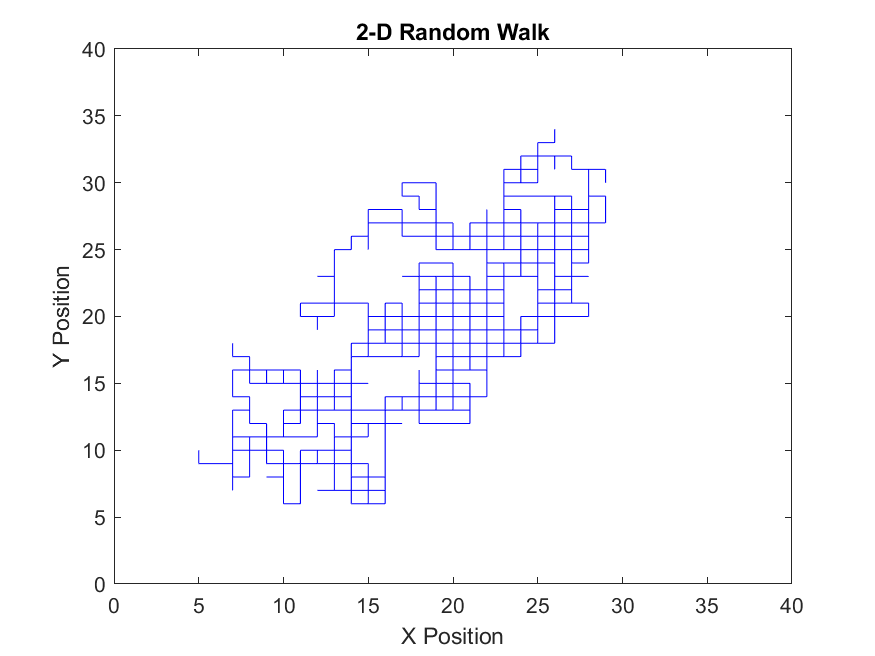} 
    \caption{Random Walk.} 
    \label{fig8:a} 
  \end{subfigure} 
  \begin{subfigure}[b]{0.45\linewidth}
    \centering
    \includegraphics[width= 1.0\linewidth]{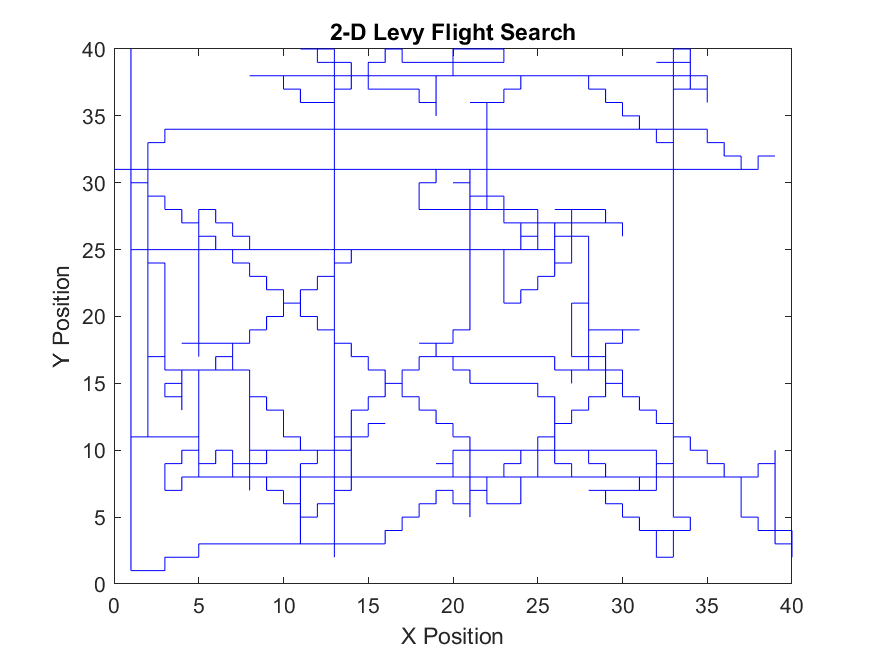}  
    \caption{L\'evy Flight Search.} 
    \label{fig8:b} 
  \end{subfigure} 
\caption{Example paths for two different rollout policies. The long lines in the L\'{e}vy flight search are artifacts from the periodic boundaries. }
  \label{fig8} 
\end{figure}

For each simulation, the Monte Carlo Tree Search will follow a default rollout policy to reach a terminal node. The goal of this policy is to give an accurate representation of how the chosen child node may affect the end result. Additionally, the rollout policy should take minimal computing power so that the simulations can be completed as quickly as possible. In this work we explore two different rollout policies: random selection and L\'evy flights. Examples of both are found in figure \ref{fig8}. 

First, the most basic rollout policy is random selection. This is exactly as it sounds: the program chooses one of its available moves completely randomly. This takes the least computing power and usually has good results, although there is a high variance. 

Second, we propose the L\'evy flight search as a more complex rollout policy. This method mimics the way some animals search for food in their environment, utilizing a few large jumps within a set of many small steps \cite{DynamicExtending}. The direction is chosen randomly but the distance is drawn from a stable distribution \cite{VisInRandomSearches}: 
\begin{equation}
P(l_j) \sim l_j^{-\mu} \; \; 1 < \mu \leq 3.
\end{equation}

Although Monte Carlo Tree Search is most often applied to games, its foundation is in decision theory. It is beginning to be applied to a much wider array of problems \cite{mandziuk2018}.  A similar application of MCTS to the search detection scenario described above is called the multi-armed bandit problem \cite{auer2002,kocsis2006}.  There a general game is played where, given a node $v$, many different arms can be played. The resulting play ends in a reward between zero and 1. In a follow on paper \cite{kocsis2006improved}, a rigorous proof was given with restrictive assumption on the game and the rewards. 

In this work we show that MCTS is an efficient method given various amounts of information. For example, given total information about the target (i.e. it is drawn from a delta distribution), the searcher finds an almost optimal path given a finite decision time. As the information is more uncertain (the distribution of the target widens into a Normal distribution) MCTS out performs, in number of steps to find the target, any default search policy such as a random walk, L\'{e}vy flight search, and nearly self avoiding random walk. The proof for the delta distributed target could be done by showing that the reward function $Q$ satisfies the assumptions stated in \cite{kocsis2006improved}. However the proof done here is streamlined and shows insight to which parts of the default policy are key. Namely that the default policy must have a stopping time which, on average, is shorter the closer you start to the target. For the unknown target case the theorem and proof here are novel. They show that the nearly self avoiding random walk, or searching only in sites that have not been searched before, is the optimal case when there is no information about the target. 

The outline of the paper is as follows. In section \ref{sec:Meth} the search algorithm is defined for the UCT as well as the target distribution. In section \ref{sec:Sims}, we present simulation data for various cases of information on the target. From complete information, delta distribution target, to fully unknown target, uniformly distributed target. In section \ref{sec:Theo} the theorems for convergence to the optimal path for the delta distribution target, and convergence to a nearly self avoiding random walk for the uniformly distributed target are stated and proved. In section \ref{sec:Conclusion} the results are summarized. 

\section{Methodology}
\label{sec:Meth}
For each target distribution the MATLAB\texttrademark \; program follow a similar setup \cite{github}. First, we set the parameters such as grid size ($N$), number of trials ($T$), number of loops/simulations ($L$), vision radius ($r_v=1$) and exploration constant ($c$). Of note, the vision radius is the maximum distance between searcher and target which registers as a detection. The exploration constant is a parameter of the UCT algorithm which balances the tendency to exploit vs. explore the decision space \cite{BrowneSurvey}. Next, we place the target in the search space according to the desired distribution. 

There are three general cases of this search problem based on the information known about the target. We can either have a completely known target drawn from a delta distribution, a completely unknown target drawn from the uniformly random distribution, or a somewhat known target drawn from the Gaussian distribution with some specified value of standard deviation, $\sigma$. For this paper, we use the Gaussian distribution as the general target distribution and we can be changed by varying the standard deviation ($\sigma$). The target distribution is defined
as follows. Let $(x,y)\in\mathbb{R}^2$ be normally distributed random variables with mean $N/2$ and standard deviation $\sigma$ for both $x$ and $y$. Then the target is selected by 
\begin{align}
	(x,y)&\sim \mathcal{N}(N/2,\sigma) \\
	T = (T_x,T_y) &=  \left (\min_{i,j\in\mathbb{Z}}\{|i-x|^2+|j-y|^2\}\right ) \mod (N,N) + (1,1)
\end{align}
where the modulo operator is done for each component. 
The searcher is started at $(1,1)$ for all trials. 

The Monte Carlo Tree Search section of the code is repeated until the searcher is within $r_v$ of the target. Before each move, the program plays out $L$ loops. In each loop a practice target is placed according to the known distribution of the real target. The UCT policy is used to select a node, then the rollout policy plays the game to completion. The number of steps taken in each loop is recorded and added to the data for each searcher location that the simulation visited. After all the loops are completed the searcher will choose the best move and update its location. 

After each trial we record the total number of steps taken and compare that to the optimal steps from searcher to target. For an accurate comparison, we use the average steps over optimal (ASOO) as the final metric.

\begin{figure}[htbp] \centering 
  \begin{tabular}{cc}
    \includegraphics[width=6cm]{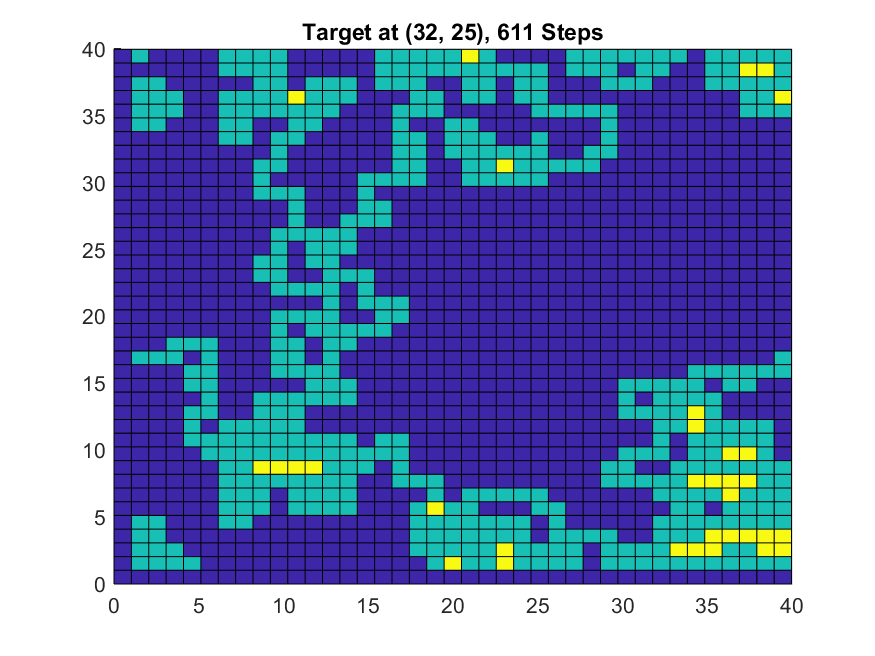}  \\
  \end{tabular}
  \caption{Example step counts for a realization of a nearly self-avoiding random walk.} 
  \label{fig:2}
\end{figure}

Throughout this work we compare the MCTS results to those of a random walk (RW), a nearly self-avoiding random walk (NSARW), and a simple L\'evy Flight Search. Both the RW and LFS are described in section 1.2. The NSARW is similar to a random walk except that it avoids previously visited locations. It is also called a weakly (or restricted) self avoiding random walk and has been studied extensively \cite{shuler2009,bauerschmidt2012}. Here, the NSARW means that each directional step is chosen randomly out of the options that have not yet been visited. If all of the targets have been visited at least once, then the direction will be chosen randomly out of the options visited the minimum number of times. An example of the NSARW path is seen in figure \ref{fig:2}, which shows the step counts for each location in an example nearly self-avoiding random walk. 

\newpage
\section{Simulation Results}
\label{sec:Sims}
\begin{figure}[htbp] 
\centering
  \begin{subfigure}[b]{0.4\linewidth}
    \centering
    \includegraphics[width=.8\linewidth]{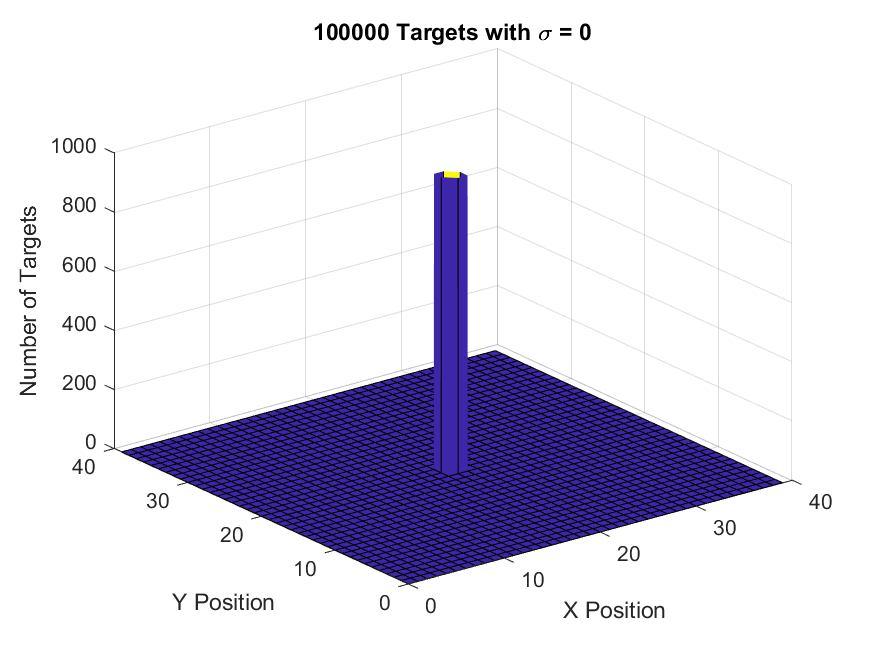} 
    \caption{Delta Distribution.} 
    \label{fig4:a} 
  \end{subfigure} 
  \begin{subfigure}[b]{0.4\linewidth}
    \centering
    \includegraphics[width= .8\linewidth]{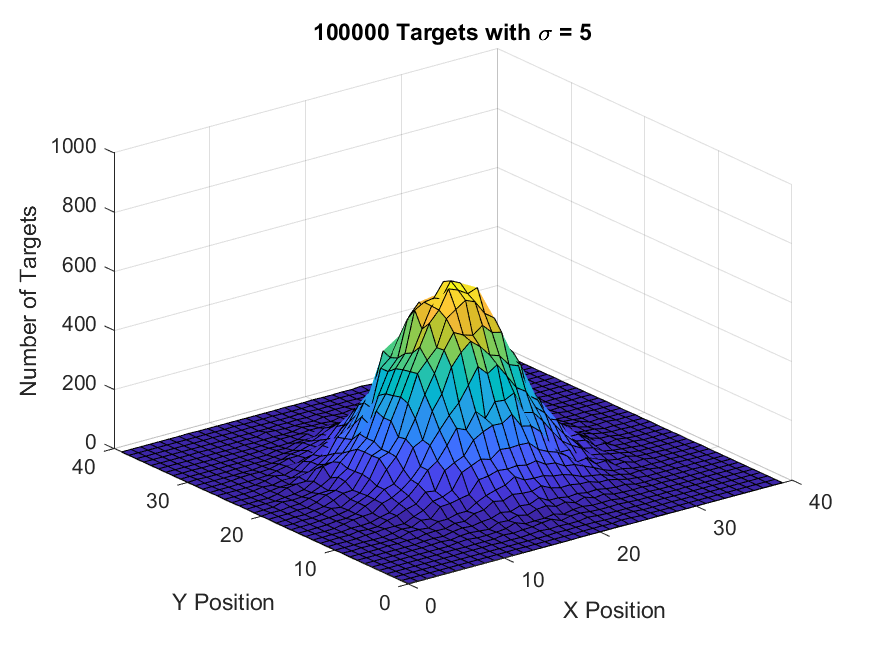}  
    \caption{Small $\sigma$ example distribution.} 
    \label{fig4:b} 
  \end{subfigure} 
  \begin{subfigure}[b]{0.4\linewidth}
    \centering
    \includegraphics[width= .8\linewidth]{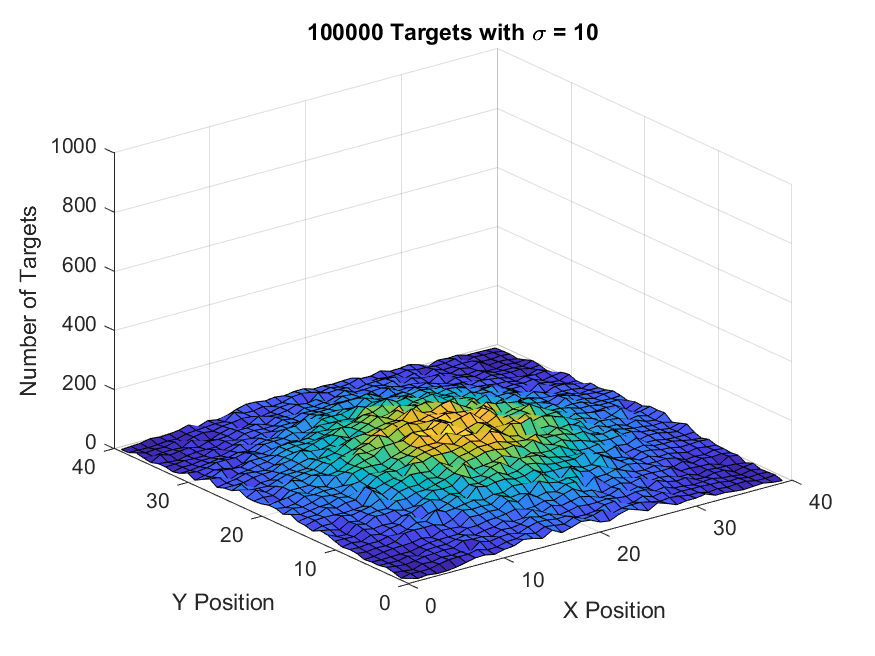}  
    \caption{Large $\sigma$ example distribution.} 
    \label{fig4:b} 
  \end{subfigure} 
  \begin{subfigure}[b]{0.4\linewidth}
    \centering
    \includegraphics[width= .8\linewidth]{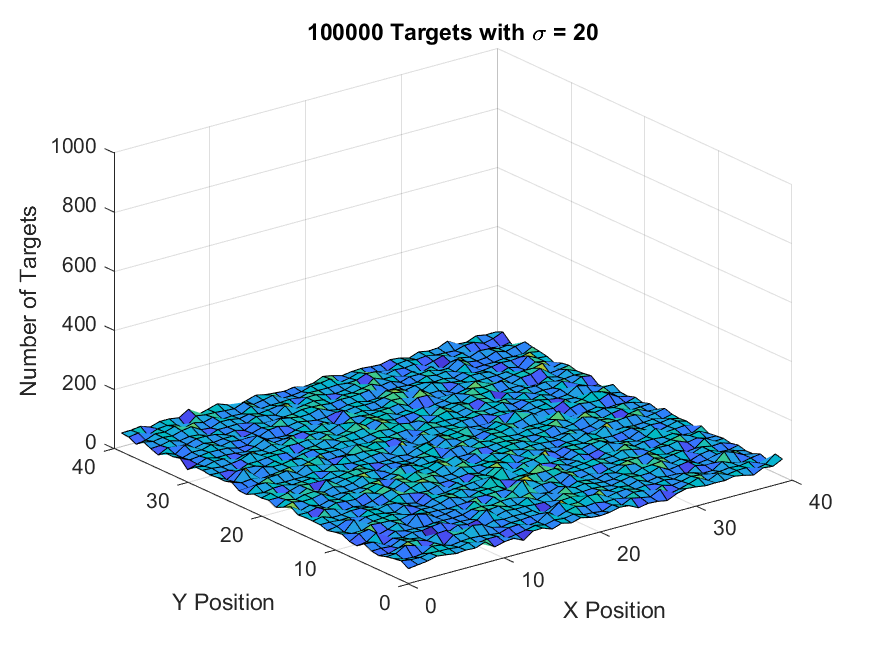}  
    \caption{Uniform distribution.} 
    \label{fig:Gauss} 
  \end{subfigure} 
\caption{Example Gaussian target histogram with various $\sigma$ values. Each histogram is run with $10^4$ targets and bins at each 
lattice point $(i,j)\in\mathbb{Z}^2$.}
  \label{fig4} 
\end{figure}

To simulate various cases we vary the standard deviation, $\sigma$ of the target distribution. We can view this as a range, where $\sigma = 0$ corresponds to the known target (delta distribution) case and $\sigma = \infty$ corresponds to an unknown target (uniformly random distribution). These distributions are shown in figure \ref{fig:Gauss} for four different values of $\sigma$. Accordingly, the simulated data shows that as $\sigma$ increases, our performance decreases, leveling out to match the unknown target case (a nearly self-avoiding random walk). This data is shown in figure \ref{fig:Gauss}. It is also important to note that for any value of $\sigma$ the Monte Carlo Tree Search with random walk default policy outperforms both a basic L\'evy Flight Search (LFS) and a random walk (RW).   \\

\begin{figure}[htbp] \centering 
  \begin{tabular}{cc}
    \includegraphics[width=7cm]{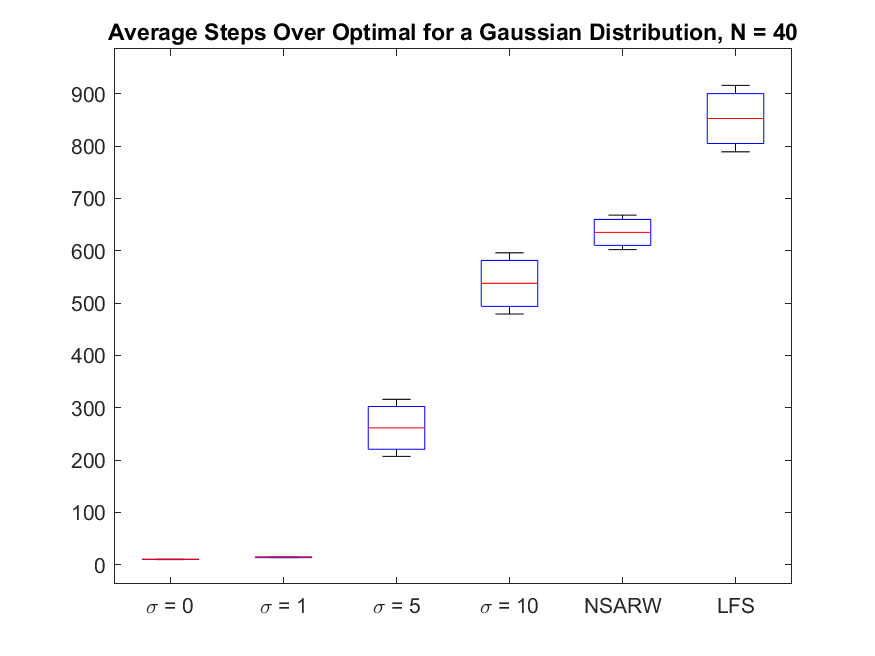} 
  \end{tabular}
  \caption{Performance with a Gaussian Target Distribution, 95\% Confidence Interval. Each box and whisker is taken over $10^3$ trials. } 
  \label{fig:GaussResults}
\end{figure}

\subsection{Special Case: Known Target}
\begin{figure}[htbp] \centering
  \begin{tabular}{cc}
    \includegraphics[width=10cm]{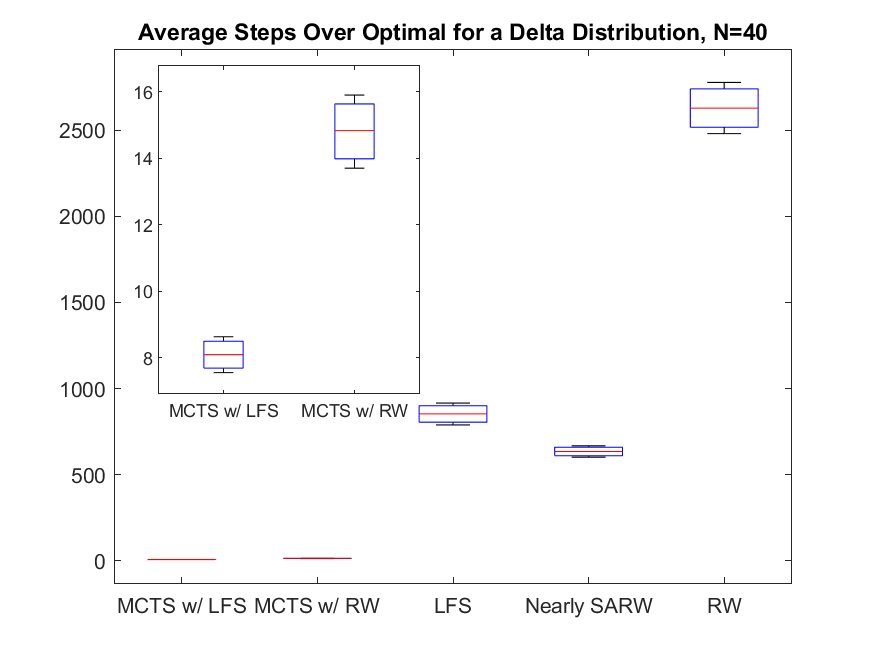}  \\
  \end{tabular}
  \caption{Performance with a Delta Target Distribution, 95\% Confidence Interval. Averaged over $10^3$ trials. }
  \label{fig:3}
\end{figure}

One special case of the Gaussian distribution occurs when $\sigma = 0$. This is known as the delta distribution or in our case, the known target scenario. Since the standard deviation is $0$, every time a target is drawn it will be in the same spot. This means that the simulated trials will be a very useful representation of the real problem, resulting in a very efficient search method. Figure \ref{fig:3} shows how the MCTS compares to other methods under these conditions. It is interesting to see that the L\'evy Flight Search default policy results in a more efficient search than the random walk policy. We hypothesize that this is a result of the LFS covering more of the search space and back-tracking across its path less. 

In section \ref{sec:Optimal} we will prove that in this scenario the MCTS converges to the optimal path. Supporting data is shown in figure \ref{fig4} with both the number of loops per step and the decision time as independent variables. For the decision time case (figure \ref{fig4} (a)) the random walk out performs the Levy Flight Search because it is faster at finding targets in moderately sized grids. 

\begin{figure}[htbp] 
\centering
  \begin{subfigure}[b]{0.45\linewidth}
    \centering
    \includegraphics[width=1.0\linewidth]{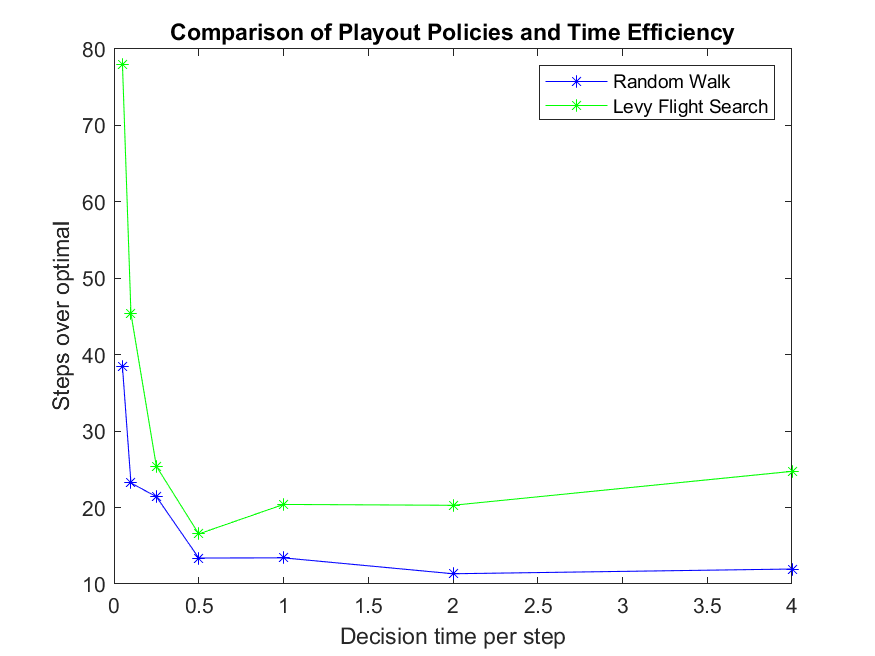} 
    \caption{Effect of decision time for MCTS. } 
    \label{fig4:a} 
  \end{subfigure} 
  \begin{subfigure}[b]{0.45\linewidth}
    \centering
    \includegraphics[width= 1.0\linewidth]{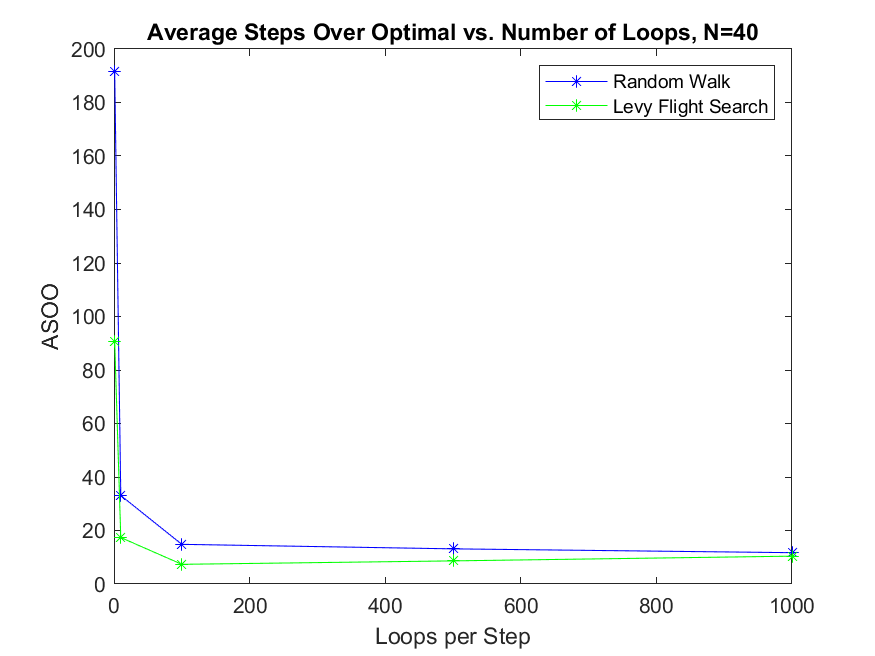}  
    \caption{Effect of loops for MCTS.} 
    \label{fig4:b} 
  \end{subfigure} 
\caption{Effect of resource limits on MCTS with each default policy. Each data point is the average over $10^3$ trials. }
  \label{fig4} 
\end{figure}

\subsection{Special Case: Unknown Target}
\begin{figure}[htbp]\centering
  \begin{tabular}{cc}
    \includegraphics[width=9cm]{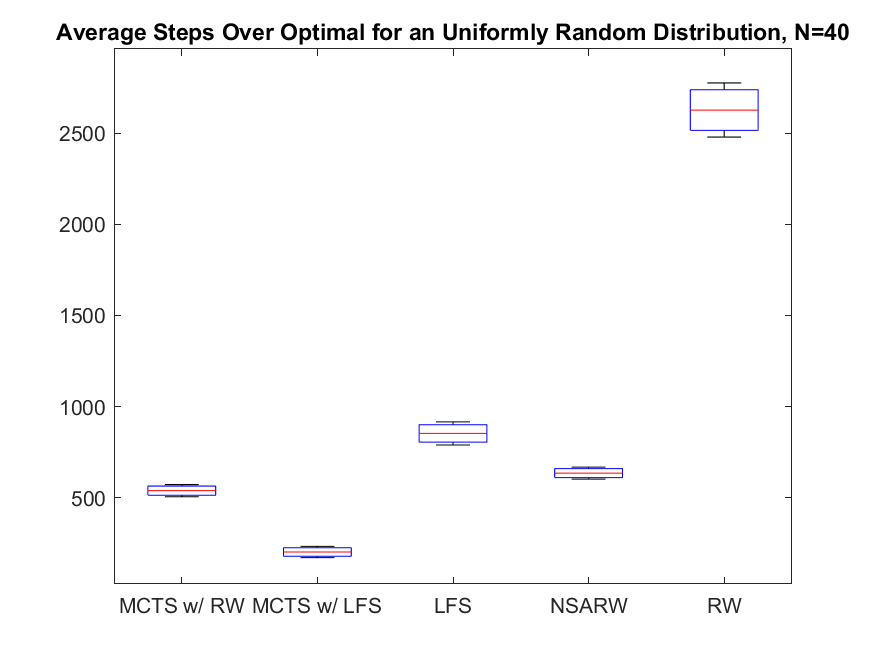}  \\
  \end{tabular}
  \caption{Performance with a Uniform Target Distribution, 95\% Confidence Interval over $10^3$ trials.} 
  \label{fig:5}
\end{figure}

When $\sigma$ is large enough for the given search space the Gaussian distribution functions as a uniformly random distribution. This is the most difficult scenario for the searcher to find an efficient path because it knows nothing useful about the target's location. Nevertheless, for both LFS and random walk default policies the MCTS does outperform a basic LFS and a basic random walk. The supporting data is shown in figure \ref{fig:5}.

\begin{figure}\centering 
  \begin{tabular}{cc}
    \includegraphics[width=8cm]{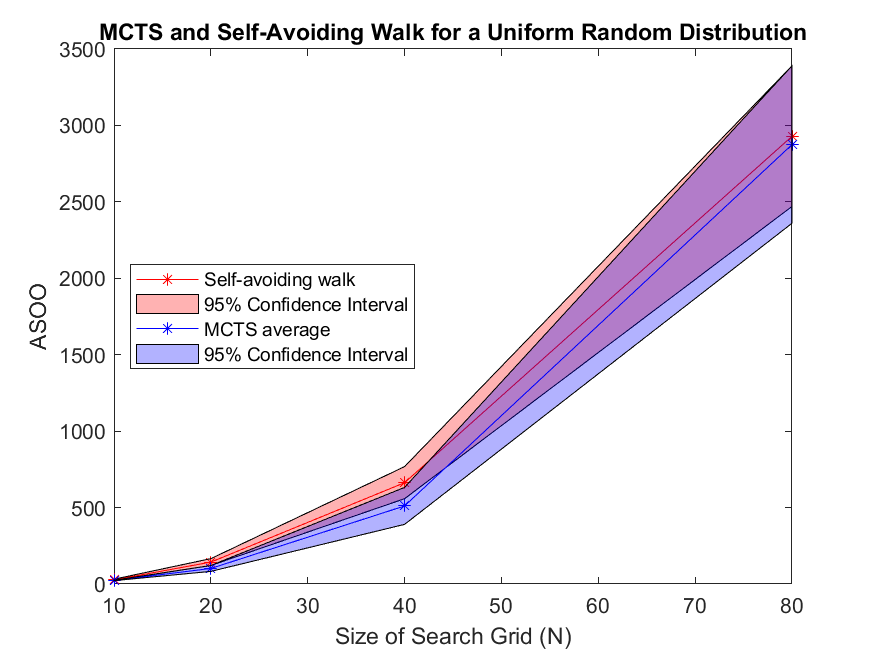}  \\
  \end{tabular}
  \caption{Convergence of MCTS to a Self-Avoiding Random Walk. Each point is the average of $10^3$ trials. The confidence interval is linearly approximated for values between data points. } 
  \label{fig:6}
\end{figure}

Additionally, we show in figure \ref{fig:6} that the MCTS with random walk default policy will converge to a nearly self-avoiding random walk as the size of the search grid gets larger, assuming the target is drawn from a uniformly random distribution. The nearly self-avoiding random walk (NSARW) is defined as a path where the searcher picks its move randomly but avoids previously visited spaces unless there are no other options. We provide a proof of this in section \ref{sec:SAW}. 

\section{Theoretical Results}
\label{sec:Theo}
Here we provide two proofs regarding the behavior of a searcher using the MCTS method to find a single target. The results are dependent on the knowledge of the target distribution and are supported by the data above. 

\subsection{Convergence to the Optimal Path}
\label{sec:Optimal}
\textbf{Assumptions:} We start with a finite search lattice of size $N \times N$. A target is placed randomly within, at location $T=(T_x, T_y)$. The searcher is started from $(1,1)$. An optimal path is defined as a series of directional moves that results in the searcher reaching the target in as few time steps as possible. The optimal number of steps is calculated by using the $\ell^1(\mathbb{T}^2)$ metric on a torus 
$$\|((1,1)-(T_x,t_y)) \|_{\ell^1}=\left  \{\begin{array}{ll}
	\Big|1 - T_x \Big| + \Big|1- T_y \Big|& \mbox{ if } \Big|1 - T_x \Big|\leq N/2,\quad  \Big|1- T_y \Big|\leq N/2, \\
	N/2 -\Big|1 - T_x \Big| + \Big|1- T_y \Big|, & \mbox{ if } \Big|1 - T_x \Big|>N/2, \quad \Big|1- T_y \Big|<N/2\\
	 \Big|1 - T_x \Big|+ N/2-\Big|1- T_y \Big|, & \mbox{ if } \Big|1 - T_x \Big|<N/2, \quad \Big|1- T_y \Big|>N/2 \\
	N -\Big|1 - T_x \Big| - \Big|1- T_y \Big|, & \mbox{ if } \Big|1 - T_x \Big|>N/2, \quad \Big|1- T_y \Big|>N/2 
	\end{array} \right .$$
where the $\mathbb{T}^2$ is suppressed for the remainder of the paper. Finally we assume that the default policy chosen has a finite stopping time $E[\tau_D,N]$ of finding the target starting $D$ steps away on an $N\times N$ grid and 
$$E[\tau_D,N]<E[\tau_{D+2},N].$$

\textbf{Theorem 1:} For a target with the delta distribution, as the number of simulation loops goes to infinity, the algorithm will choose the move that leads the searcher in an optimal direction (i.e., choose the direction that minimizes the $\ell^1$ distance to the target).

Before proving the theorem, we use a lemma about the number of times the optimal and suboptimal choices are simulated. 
\textbf{Lemma 1:} As $l \to \infty$ for all moves $i \in \{1, ..., 4\}$ the number of times move $i$ has been played, $T_i(l) \to \infty$. \\
\begin{proof} We proceed by contradiction. Assume at lease one move (denote as $j$) is chosen a finite number of times, i.e., $T_j(l)$ is bounded. Thus we can find some constant, $T$ such that for all $l$, $$T_j(l) \leq T.$$
Based on the UCT method, the program will chose its next move by maximizing 
$$UCT(i) = Y_{i, T_i(l)} + c \sqrt{\frac{\log{l}}{T_i(l)}}.$$
Using the assumption that one move is bounded, we know that at least one of the other moves must not be bounded since the total number of simulations goes to infinity.\\
Since $0 \leq Y_{i, T_i(l)} \leq 1$ we can say that 
$$UCT(j) \leq 1 + c\sqrt{\frac{\log{l}}{T_j(l)}}$$ $$UCT(j) \geq c\sqrt{\frac{\log{l}}{T}}$$
For the unbounded move, $i$, we know that as $l \to \infty$, $T_i(l) \to \infty$. For all moves, $\log{l}$ is the same. \\
We choose an $L$ such that for all $l \geq L$, $\; T_i(l) > T^2$. \\
Thus $$\frac{1}{T_i(l)} < \frac{1}{T^2} \leq \frac{1}{T}$$
Multiplying by $\log{l}$, which is the same for all arms $i$, we get $$\frac{\log{l}}{T_i(l)} \leq \frac{\log{l}}{T^2} \leq \frac{\log{l}}{T}$$
$$\sqrt{ \frac{\log{l}}{T_i(l)}} \leq \sqrt{\frac{\log{l}}{T^2}} \leq \sqrt{\frac{\log{l}}{T}}$$
Since $T$ is a constant and $\log{l}$ is monotonically increasing, we can choose $L' > L$ such that for all $l \geq L'$ we get 
$$c\sqrt{\log{l}} \Big(\frac{\sqrt{T} - 1}{T}\Big) > 1$$
$$c\sqrt{\log{l}} \Big(\frac{1}{\sqrt{T}} - \frac{1}{\sqrt{T_i(l)}}\Big) > 1$$
Thus for all $l \geq \; \max\{L, L'\}$ we conclude that  
$$UCT(i) \leq 1 + c\sqrt{\frac{\log{l}}{T_i(l)}} < c\sqrt{\frac{\log{l}}{T^2}} < c\sqrt{\frac{\log{l}}{T}} \leq UCT(j)$$
Therefore $UCT(i) \leq UCT(j)$ so move $j$ will be chosen and played out. \\
This contradicts the assumption that $T_j(l)$ is bounded, so we know that as $l \to \infty$ $T_j(l) \to \infty$. The same logic can be applied to all other moves $i \in \{1, ..., 4\}$ so we conclude that for all $i$, as $l \to \infty$ $T_i(l) \to \infty$. \\
\end{proof}


\begin{proof}
Assume the searcher is at point $S = (S_x,S_y)$. 
For the searcher choosing between $4$ possible directions ($1 = $ up, $2 = $ right, $3 = $ down, $4 = $ left) let $i \in \{1, ..., 4 \}$ represent a suboptimal choice and $i_*$ represent the optimal move. 
We define $Y_{i, T_i(l)}$ as the average reward for move $i$ after $l$ loops have been played out. Here, $T_i(l)$ is the number of times direction $i$ has been played after $l$ loops. Note that as $l \to \infty$, $T_i(l) \to \infty$ for all moves $i \in \{1, ..., 4\}$ ((Lemma 1). We calculate the average reward as 
$$ Y_{i, T_i(l-1)} = \frac{1}{X_{i, T_i(l)}}$$ where $X_{i, T_i(l)}$ represents the average step count, defined as 
$$X_{i, T_i(l)} = \frac{\tau_{1, D+1, N} + \tau_{2, D+1, N} + ... \tau_{T_i(l), D+1, N}}{T_i(l)}.$$
Each $\tau_{D, N}$ is a random variable that describes the number of steps taken in a random walk from the searcher's initial location to the target. $D$ is the number of optimal steps away from the target and $N$ is the size of the grid. Assume the searcher is exactly $D$ steps away from the target before choosing its first move. That is, 
$$ D = \|S-T\|_{\ell^1}.$$
 For a suboptimal move, $i$, the searcher initially moves one step away from the target, increasing this distance to $D+1$, hence the sequence of $\tau_{D +1, N}$. For the optimal arm, $i_*$, however, the searcher moves toward the target, decreasing the distance to $D-1$. Thus we define the average step count for $i_*$ after $l$ loops as 
$$X_{*, T_*(l)} = \frac{\tau_{1, D-1, N} + \tau_{2, D-1, N} + ... \tau_{T*(l), D-1, N}}{T_*(l)}.$$
The infinite sequence of random walks starting from the same distance converges to an expected value of $E[\tau_{D, N}]$ . Since farther distances will on average take longer to reach the target, $E[\tau_{D-1, N}] < E[\tau_{D+1, N}]$. 
By the law of large nubmers we can say that for a suboptimal move, $i$, 
$$\lim_{l \to \infty} \frac{\tau_{1, D+1, N} + \tau_{2, D+1, N} + ... \tau_{T_i(l), D+1, N}}{T_i(l)} = \lim_{l \to \infty} X_{i, T_i(l)} = E[\tau_{D+1, N}]$$ with probability $1$. Similarly, for the optimal move we have  
$$\lim_{l \to \infty} \frac{\tau_{1, D-1, N} + \tau_{2, D-1, N} + ... \tau_{T*(l), D-1, N}}{T_*(l)} = \lim_{l \to \infty} X_{*, T_*(l)} = E[\tau_{D-1, N}].$$ 
Since the average reward is the inverse of the average step count, we substitute to get 
$$\lim_{l \to \infty} Y_{i, T_i(l)} = \frac{1}{E[\tau_{D+1, N}]}, \;\;\; \lim_{l \to \infty} Y_{*, T_*(l)} = \frac{1}{E[\tau_{D-1, N}]}.$$
Set $\delta > 0$. Then, for all $\delta$ we can find $L$ such that for all $l \geq L$ we know 
$$Y_{i, T_i(l)} \geq \frac{1}{E[\tau_{D+1, N}]} - \delta, \quad Y_{i, T_i(l)} \leq \frac{1}{E[\tau_{D+1, N}]} + \delta, \quad Y_{*, T_*(l)} \geq \frac{1}{E[\tau_{D -1, N}]} - \delta, \quad Y_{*, T_*(l)} \leq \frac{1}{E[\tau_{D -1, N}]} + \delta$$

For each real move of the searcher, the MCTS method chooses the direction with the best UCT value: 
$$I_l = \max_{i \in \{1, ..., K\}} \{Y_{i, T_i(l-1)} + c_{l-1, T_i(l-1)}\}, $$ where $c_{l, s}$ is a bias sequence chosen to be $$c_{t,s} = \sqrt{\frac{2 \ln t}{s}}.$$
Fix $\epsilon > 0$. We want to show that if $l$ is sufficiently large, $\mathds{P}(I_l \neq i_*) \leq \epsilon$. 
For any suboptimal move, $i$, let $p_{i, l} = \mathds{P}(Y_{i, T_{i, T_i(l)}} \geq Y_{*, T_*(l)})$. We substitute the inverse of our average reward to get $p_{i, l} = \mathds{P}(X_{*, T_*(l)} \geq X_{i, T_i(l)})$.  Then we know $\mathds{P}(I_l \neq i_*) \leq \sum_{i \neq i_*} p_{i, l}$. Therefore it suffices to show that $p_{i, l} \leq \frac{\epsilon}{K}$ is true for all suboptimal arms when $l$ is large enough. \\
Using the limits above, if $X_{*, T_*(l)} < E[\tau_{D-1, N}] + \delta$ and $X_{i, T_i(l)} > E[\tau_{D -1, N}] - \delta$ then $X_{*, T_*(l)} < X_{i, T_i(l)}$. Therefore, 
$$p_{i, l} \leq  \mathds{P}(X_{*, T_*(l)} \geq E[\tau_{D-1, N}] + \epsilon) + \mathds{P}(X_{i, T_i(l)} \leq E[\tau_{D+1, N}] - \epsilon).$$
By the law of large numbers, as $n \to \infty$, $\mathds{P}(|E[X] - X_n| > \delta) \to 0$. We apply this to the equality above to say that 
$$\lim_{l \to \infty} p_{i, l} \leq  \lim_{l \to \infty} P(X_{*, T_*(l)} - E[\tau_{D-1, N}] \geq \epsilon) + \lim_{l \to \infty} P( E[\tau_{D+1, N}] - X_{i, T_i(l)} \geq \epsilon) = 0.$$
Then, since $\lim_{l \to \infty} p_{i, l} = 0$ we can choose an $L$ sufficiently large such that for all $l \geq L$ 
$$p_{i, l} \leq \frac{\epsilon}{K}.$$
\end{proof}

\subsection{Convergence to a Self-Avoiding Random Walk}
\label{sec:SAW}
\textbf{Theorem: } A Monte Carlo Tree Search that uses the UCT selection policy with random walk default policy and for a uniformly distributed target will converge to a nearly self-avoiding walk as $M$, the size of the search lattice, goes to infinity. \\

\begin{proof}
We want to show that the UCT algorithm, defined by $$UCT(v') = \frac{Q(v')}{N(v')}\ + c \sqrt{\frac{log(N(v))}{N(v')}},$$ chooses a child node ($v'$) randomly as the number of trials goes toward infinity, unless one direction has been picked more frequently than the others.  \\

Let $T$ be the randomly selected target location and $S_{t}$ be the location of the searcher at time $t$. Assume that $S_0 = (0,0)$. For all $\epsilon > 0$ choose $K_{\epsilon}$ such that $\frac{1}{K_{\epsilon}} < \epsilon$ as the radius of a region $E$ around the origin. Thus $K_{\epsilon}$ is the minimum number of steps required to exit this region. Then choose $M$ as the dimension of a square search grid where $T$ is located randomly within it such that $P(dist(T, S_0) > K_{\epsilon}) = 1 - \delta$ where $dist(x,y)$ returns a discrete value of minimum steps between $x$ and $y$ and $\delta$ is arbitrarily small. This is represented pictorially in figure \ref{fig:SAWproof}. 

\begin{figure}
\centering
\includegraphics[scale=1]{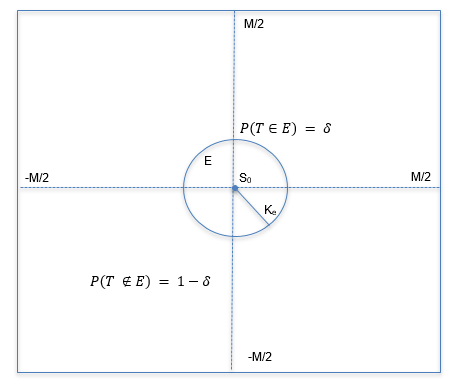}
\caption{Search grid given $\epsilon$.}
\label{fig:SAWproof}
\end{figure}

Let $v'_i, \; i\in \{1, 2, 3, 4\}$ represent each of the four child nodes (up, down, left and right) of the root node $v$. Then $Q(v')$ represents the success rate of this node, where $Q(v') = \frac{1}{k} $ where $k$ is the average number of steps for many trials. Recall that $K_{\epsilon}$ is the step count for an optimal path, so on average $k\gg K_{\epsilon}$. This means that as the number of trials increases we get $Q(v')\ll \frac{1}{K_{\epsilon}} < \epsilon$. Therefore as $\epsilon$ gets smaller, $Q(v')$ approaches zero. Hence the first term of the UCT algorithm approaches zero.

The second term, $c \sqrt{\frac{log(N(v))}{N(v')}}$, is a relationship between the visit counts of the parent and child nodes. For the first four trials, each child node gets visited once (chosen in a random order as per MCTS protocol). It is important to note that after the first step is taken, the previous root node (now a child node) will have a very large visit count because it was included in all of the simulations from the past decision. Thus, after 4 trials, $N(v') = 1 \; \; \forall \; v'_i,\; i\in \{1, 2, 3, 4\}$ except the direction from which it came, which will be significantly higher (equal to the number of loops). Note that $N(v)$ is the same for all $v'_i$ since this new root node's visit count increases after each and every trial. 

Since $N(v')$ has equal entries for all directions except one, and $N(v)$ is a constant, the UCT vector will have equal values for three moves, excluding the direction from which it came. Thus the UCT program picks one of the three directions randomly, avoiding its previous path. Call this direction $v'_1$. The next trial will be run using $v'_1$ as the first move. Regardless of the outcome, we now get $N(v') = 1 \; \: \forall v'_i \;\: i \neq 1$ (except the previously visited node), and $2$ for $v'_1$. Since the UCT value has $N(v')$ in the denominator of the second term, the UCT value will decrease for this $v'_1$ that has been visited twice. Therefore $max(UCT)$ will randomly return one of the three remaining values (these are all equal). 

The process will repeat until all four nodes are visited twice, then three times, etc. Regardless of the time this program runs it will continue randomly cycling through the nodes, trying to avoid hitting one node more times than the others. At the end, all four nodes will have an equal visit count so the final move decision will be chosen uniformly random. 

Therefore the program will choose a random direction unless one or more of the directions have been chosen an unequal number of times, in which case the program will choose randomly between the options visited the minimum number of times. Thus the path becomes a nearly self-avoiding random walk.  
\end{proof}

\section{Conclusion}
\label{sec:Conclusion}
In this paper we studied a search detection game on a 2-D finite lattice with periodic boundaries. The target was chosen using a normal distribution with varying standard deviation $\sigma$.  The searcher followed the Monte Carlo Tree 
Search algorithm with the UCT selection policy with the rewards based on the inverse of the time it takes the searcher to find the target. In the study, the standard deviation of the target was varied. Additionally two default policies for the MCTS were studied, a random walk and L\'{e}vy flight search. 

From numerical simulations, we find the MCTS learns to find the target efficiently for all cases of $\sigma$. Although there are diminishing returns for $\sigma>10$ which corresponds to 95\% of the targets being found within a $20\times 20$ subsection of the $40\times 40$ grid. In the case where the target is known (drawn from the delta distribution), the searcher converges to the optimal bath as the number of simulation loops (or decision time) grows. Interestingly, we found that between the two default policies, L\'{e}vy flight search out performs random walk based on loops per decision where as the random walk default policy is better for a restricted decision time. This could be because the random walk is more efficient at searching the moderate sized grid ($40\times 40$) that we were using. We expect that a L\'{e}vy flight default policy would be optimal in both cases on larger grids. 

Theoretically we prove two theorems for this game. The first theorem shows that MCTS will converge to the optimal path under mild assumptions. The key to the proof are that the default policy must have a stopping time which, on average, is shorter the closer you start to the target.  The second theorem shows that for a uniformly selected target, i.e. no information given, as the grid grows large the MCTS with the random walk default policy converges to a nearly (or weakly) self avoiding random walk. Intuitively this makes sense. The searcher only looks for the target in previously unvisited sites. 

\bigskip 

\textbf{Acknowledgements:}  The work of S. Hottovy is partially supported by the National Science Foundation under Grant DMS-1815061. E. Kozak was
supported as a student by the NSF grant. This project is also part of the Trident Scholar program at USNA and received support and feedback from the Trident committee, especially in the School of Math and Science.

\bibliographystyle{plain}
\bibliography{refs}

\end{document}